\documentclass[10pt,twocolumn,letterpaper]{article}

\usepackage{cvpr}
\usepackage{times}
\usepackage{epsfig}
\usepackage{graphicx}
\usepackage{amsmath}
\usepackage{amssymb}
\usepackage{upgreek}
\usepackage{cite}
\usepackage{algorithm}
\usepackage{algorithmic}
\usepackage{wrapfig}
\usepackage{subcaption}

\usepackage[pagebackref=true,breaklinks=true,letterpaper=true,colorlinks,bookmarks=false]{hyperref}

\cvprfinalcopy 

\def\cvprPaperID{4294} 

\newcommand{\john}[1]{\textcolor{red}{#1}}

\ifcvprfinal\fi
\begin{document}
\title{
Exploit Clues from Views: \\
Self-Supervised and Regularized Learning for Multiview Object Recognition\\
}

\author{Chih-Hui Ho \qquad Bo Liu \qquad Tz-Ying Wu \qquad Nuno Vasconcelos\\
University of California, San Diego\\
{\tt\small \{chh279,boliu,tzw001,nvasconcelos\}@ucsd.edu}}

\newcommand{\newName}{\john{newName}}
\newcommand\given[1][]{\:#1\vert\:}
\maketitle

\begin{abstract}
Multiview recognition has been well studied in the literature and achieves decent performance in object recognition and retrieval task. However, most previous works rely on supervised learning and some impractical underlying assumptions, such as the availability of all views in training and inference time. In this work, the problem of multiview self-supervised learning (MV-SSL) is investigated, where only image to object association is given. Given this setup, a novel surrogate task for self-supervised learning is proposed by pursuing ``object invariant" representation. This is solved by randomly selecting an image feature of an object as object prototype, accompanied with multiview consistency regularization, which results in view invariant stochastic prototype embedding (VISPE). Experiments shows that the recognition and retrieval results using VISPE outperform that of other self-supervised learning methods on seen and unseen data. VISPE can also be applied to semi-supervised scenario and demonstrates robust performance with limited data available. Code is available at \url{https://github.com/chihhuiho/VISPE}
\end{abstract}

\section{Introduction}
\begin{figure}
    \centering
    \includegraphics[width=\linewidth]{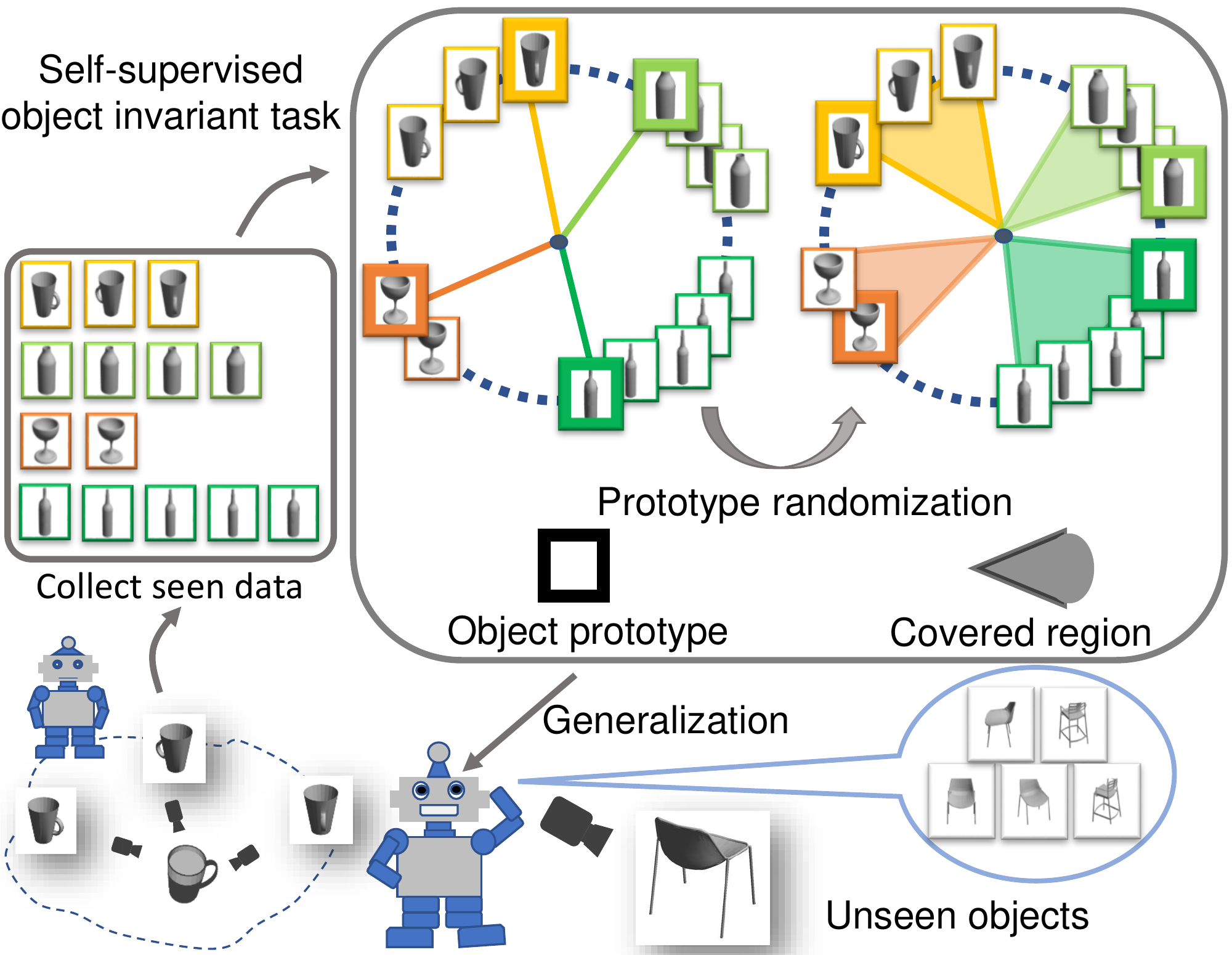}
    \caption{Lightweight unsupervised multiview object recognition. A household robot collects multiple object views by moving around, aggregating a multiview object database without view labels. A self-supervised learning algorithm is applied to this database to create an embedding that maps images from same object into an object invariant. At inference time, this embedding generalizes to new views, objects, and object classes.  }
    \label{fig:intro_fig}
    \vspace{-10pt}
\end{figure}

3D recognition has received increasing attention in computer vision in recent years. A popular approach, which we pursue in this work, is to rely on the multiview object representation. Several multiview recognition approaches have been proposed in the literature, including the use of recurrent neural networks\cite{3D2SeqViews, SiameseBiLSTM, MVLSTM}, feature aggregation from different views\cite{AAAI1817062,GVCNN,MVCNN}, graph modeling\cite{Hypergraph}
and integration with other modalities\cite{PVNet,HegdeZ16,unicode, MVvoxel}. 
While achieving good recognition performance, two strong assumptions are made. The first is that 
a dense set of views, covering the entire range of view angles, is available per object \cite{MVCNN}. While some methods support missing views during inference \cite{Ho_2019_CVPR,Kanezaki16,GVCNN}, a complete view set is always assumed for training. The second is that all these images are labeled, for both object classes and view angle. The two assumptions make multiview techniques difficult to implement and limit their generalization. For example, while previous works~\cite{MVCNN,Kanezaki16,GVCNN} show strong performance on training classes, recognition on classes unseen during training is usually not considered.

These limitations prevent many applications of interest. Consider, for example, the setting of Fig.~\ref{fig:intro_fig}, where a household robot of limited memory is tasked with
picking scattered objects and returning them to their locations. In this setting, it is impractical to pre-train the robot with a dense set of labelled views for each object class in the world. Instead, the robot must be able to efficiently learn 
objects from unseen classes
after deployment.
This is similar to problems like image retrieval~\cite{image_retrieval,image_retrieval2} or face verification~\cite{FaceNet,NormFace}, which are usually solved by metric learning. 
An embedding is learned from a large dataset of annotated objects, unseen object classes are modelled by projecting example images onto the embedding, and classification is performed with a nearest neighbor classifier. 
However, a multi-view embedding is challenging to learn in this manner, due to the need for complete and labeled sets of views. In the setting of Fig.~\ref{fig:intro_fig} this means that, after the home robot is deployed, view angle labels must be collected by manually controlling the pose of the training objects, which is impractical. 

This problem can be avoided by the introduction of {\it multiview self-supervised learning\/} (MV-SSL) methods. SSL is now well established for problems where annotation is difficult~\cite{Jiang18,self_supervised_survey}. The idea is to use ``free labels," i.e. annotations that can be obtained without effort, to define a surrogate learning task. However, the many surrogate tasks proposed in the literature~\cite{egomotion,pathakCVPR16context,Hsin17,UEL,coloring,LarssonMS16} are poorly suited for multiview recognition. This is because multiview embeddings must enforce an {\it invariance\/} constraint, namely that all views of an object map into (or cluster around) a single point in the embedding, which is denoted the {\it object invariant\/}. 
For embeddings with this property, views of objects unseen during training will naturally cluster around object invariants, without requiring view labels, consistency of view angles across objects, or even the same number of views per object.  In this case, it suffices for the home robot to collect a set of views per object, e.g. by moving around it, as illustrated in Fig.~\ref{fig:intro_fig}. To emphasize the  low-complexity of object acquisition under this set-up, we refer to it as {\it lightweight unsupervised multiview object recognition\/} (LWUMOR).


In this work, we seek embeddings with good LWUMOR performance. We consider proxy embeddings~\cite{proxynca}, which have been shown to perform well for multiview recognition when dense views and class labels are available~\cite{Ho_2019_CVPR}. To derive an SSL extension, we propose a new surrogate task, where object instances are used as training ``classes,"
i.e. object identities serve as free labels for learning.
We hypothesize, however, that due to the concentration of supervision on class prototypes, these embeddings \textit{only} capture the metric structure of images in the neighborhood of these prototypes, thus overfitting to the training classes. 
We address this problem with a randomizing procedure, where the parameters of the softmax layer are sampled stochastically from the embeddings of different object views, during training. This has two interesting consequences. First, it forces the learning algorithm to produce an embedding that supports many classifiers, spreading class supervision throughout a much larger region of feature space, and enhancing generalization  beyond the training classes. Second, because this supervision is derived from randomized object views, it encourages a stable multiview representation, even when only different view subsets are available per object.

To further enhance multiview recognition performance, this randomization is complemented by 
an explicit invariance constraint, which encourages the classifier parameters to remain stable under changes of view-point. We denote the resulting MV-SSL embeddings as {\it
view invariant
 stochastic prototype embeddings (VISPE)\/}.
Experimental results on popular 3D recognition datasets show that self-supervised VISPE embeddings combine 1) better performance outside the training set than standard classification embeddings, and 2) faster convergence than metric learning embeddings.
Furthermore, for multiview recognition, VISPE embeddings outperform previous SSL methods.

Overall, this work makes three main contributions. The first is the LWUMOR formulation of MV-SSL. This enables multiview recognition without object class or pose labels, and generalizes well to objects unseen at training time. The second is a new surrogate task 
that relies on randomization of object views to encourage stable multiview embeddings, and outperforms previous SSL surrogates for multitview recognition.  The third is the combination of randomization and invariance constraints implemented by VISPE to learn embeddings of good LWUMOR performance. Extensive experiments validate the ability of these embeddings to learn good
invariants for multiview recognition.

\section{Related work}\label{sec:related_work}
This work is related to multiview recognition, SSL, and regularization by network randomization. 

\subsection{Multiview recognition}
Multiview recognition is a 2D image-based approach to 3D object recognition. One of the earliest methods is the multiview CNN (MVCNN)\cite{MVCNN}, which takes multiple images of an object as input and performs view aggregation in the feature space to obtain a shape embedding. Representing 3D objects by 2D images has been shown effective for classification\cite{Kanezaki16,GVCNN} and retrieval\cite{TC,AngularTC}. Subsequent research extended the idea by performing hierarchical view aggregation\cite{AAAI1817062,GVCNN}. 
However, because view aggregation disregards the available supervision for neighboring relationships between views~\cite{3D2SeqViews}, recurrent neural networks~\cite{3D2SeqViews, SiameseBiLSTM, MVLSTM} and graph convolutional neural networks~\cite{Hypergraph} have been proposed to model multiview sequences. 
Aside from multiview modeling, \cite{Kanezaki16} treats viewpoint as a latent variable during optimization and achieves better classification accuracy and pose estimation. In the retrieval setting, \cite{AngularTC,TC} combine the center loss~\cite{Wen2016ADF} with a triplet loss~\cite{FaceNet} to form compact clusters for features from the same object class.

All these methods share several assumptions that make them impractical for LWUMOR. The MVCNN\cite{MVCNN} assumes that all object views are presented at both training and inference. Methods that model view sequences \cite{3D2SeqViews, SiameseBiLSTM} require even more detailed viewpoint supervision.
Previous works\cite{GVCNN,Kanezaki16,Ho_2019_CVPR} found that these methods experience a significant performance drop when only partial views are available for inference. \cite{Kanezaki16} minimized this drop by treating viewpoint as an intermediate variable, while \cite{Ho_2019_CVPR} proposed to overcome it with hierarchical multiview embeddings. All these methods assume a full set of training views.

In this work, we relax this constraint, investigating the LWUMOR setting, where only partial object views are available for both training and inference, and no view or image class labels are given. This forces the use of SSL techniques to learn the ``implicit" shape information present in a set of object views, and encourages embeddings that generalize better to unseen classes. 

\subsection{Self-supervised Learning}
SSL leverages free labels for a surrogate task to train a  deep network. Many surrogate tasks have been proposed in the literature. While we provide a brief review of many of these in what follows, most do not seek object invariants and are unsuitable for MV-SSL.

\noindent\textbf{Context} based approaches \cite{pathakCVPR16context,Doersch2015UnsupervisedVR,Mundhenk17} seek to reconstruct images. Autoencoders\cite{autoencoder} map images to a low dimensional latent space, from which they can be reconstructed. Similarly, a context encoder\cite{pathakCVPR16context} reconstructs missing patches from an image conditioned on their surroundings. 
\cite{Doersch2015UnsupervisedVR} further leverages spatial image context by predicting the relative positions of randomly cropped patches. Image coloring techniques, which recover the colors of grey scale images \cite{coloring} or predict pixelwise hue and chromatic distributions\cite{LarssonMS16,LarssonMS17} leverage color as a form of image context.

\noindent\textbf{Motion} based approaches \cite{egomotion,Jayaraman2017, WangG15a, PathakGDDH16}
exploit the spatiotemporal coherence of images captured by a moving agent, in terms of relative position\cite{egomotion}, optical flow\cite{PathakGDDH16} or temporal video structure \cite{WangG15a,JayaramanG15a}. The surrogate task becomes to predict camera transformations\cite{egomotion} or segmenting objects\cite{PathakGDDH16}.  

\noindent\textbf{Sequence} sorting is another popular task, where sequences can consist of randomly cropped image patches \cite{puzzle} and video clips\cite{VideoJigsaw, MisraZH16}. 
Similarly, there have been proposals to remove some color channels from image patchs\cite{damagedPuzzle} or adding various types of jitter to video clips\cite{Hsin17,Junyan19,Jiangliu19}. 

\noindent\textbf{Data augmentation} type of tasks transform images, leveraging the difference after transformation to define surrogate tasks. \cite{predRot} predicts the rotation angle of the transformed image, while \cite{UEL} learns a transformation invariant feature.

\noindent\textbf{View} based tasks have been proposed for multiple applications, such as object recognition\cite{shapecode}, hand\cite{Wan_2019_CVPR} and human\cite{Muhammed19} pose estimation. Our work is similar to \cite{shapecode} as both consider object recognition. However, \cite{shapecode} requires image sequences
while our approach has no such constraint and tackles the problem in an entirely different manner .

\noindent\textbf{Cluster} based methods group data with visual similarities into clusters and discriminate different clusters. While \cite{caron2018deep,CliqueCNNDU} group multiple images into the same cluster, \cite{exemplarCNN,wu2018}  treat each image as a cluster. Our work shares the high level idea of the latter, by treating each object as a cluster, but differs in terms of memory usage and efficiency. While \cite{exemplarCNN} is known to be computational demanding, \cite{wu2018} is both memory expensive and inefficient, by requiring storage of features from {\it all\/} dataset instances. Furthermore, each feature is updated only once per epoch, which leads to noisy optimization. Our method leverages multiple object views to avoid these problems and is more suitable for the LWUMOR setting.


%

\begin{figure*}
    \centering
    \includegraphics[width=\linewidth]{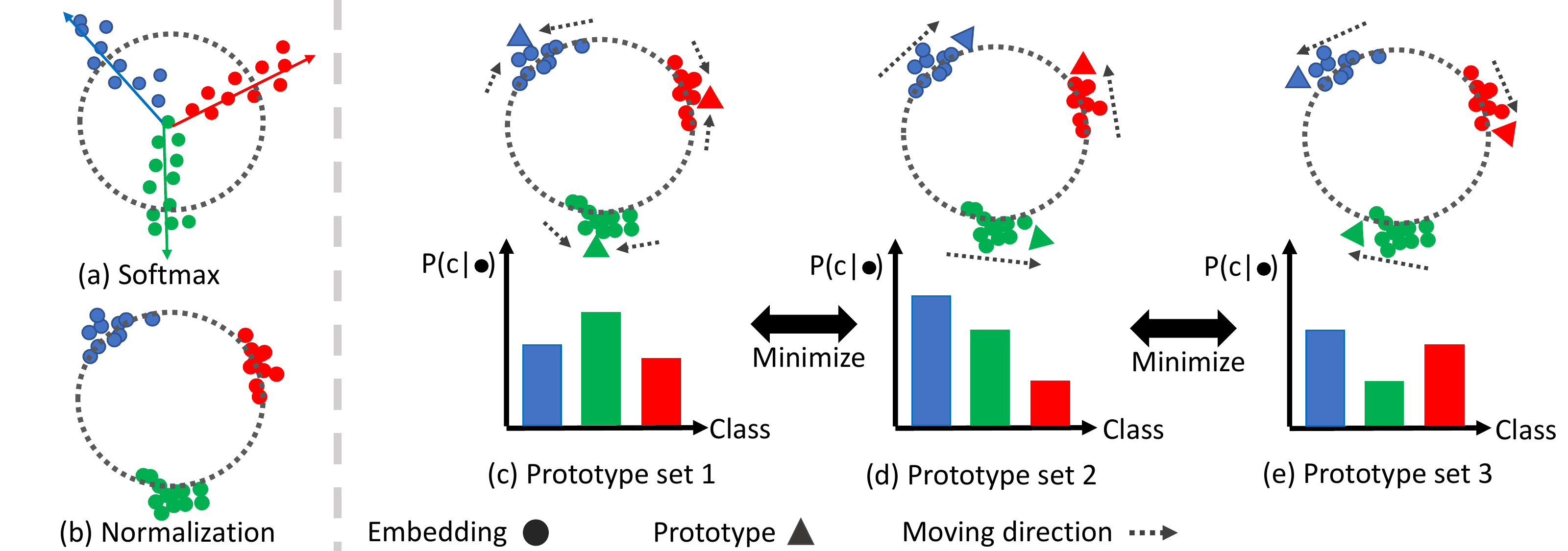}
    \caption{Regularization of a self-supervised embedding by prototype randomization. In all figures, each color represents a single object and views of the same object are marked with same color. (a) softmax embedding, unnormalized features. Solid arrows represent the weight vectors $w_i$ learned per instance $i$.  (b) Normalized embedding. (c-e) randomization: 3 different sets of prototypes are used for training. Dashed lines show how view embeddings are encouraged to move towards the corresponding object prototype. Bar plots illustrate how the posterior class probabilities of a given image change when the prototyopes are switched. The proposed multiview consistency regularization seeks to further improve the generalization power of the embedding by minimizing these variations. }
    \label{fig:rand}
    \vspace{-7pt}
\end{figure*}

\subsection{Network randomization}
Randomization has been shown to improve network performance\cite{Dropout} and robustness\cite{Cihang17}. It can be explained as a form or model ensembling \cite{NIPS2017_7219,DropMax}, by combining models trained under different conditions. One of the simplest yet most practical randomization procedures is dropout\cite{Dropout,AdaptiveDropout}, which removes units in the network during training. Dropmax\cite{DropMax} proposed to instead remove classes, training a stochastic variant of the softmax for better classification. The proposed method explores an orthogonal randomization direction, where feature vectors from different object views are chosen as object prototypes during training.

\section{Multiview Self Supervised Learning}
In this section, we discuss the proposed MV-SSL approach.

\subsection{Light Weight Unsupervised Multiview Object Recognition}

We start by defining the LWUMOR problem and introducing a surrogate task for its solution.
Consider a set of objects $\mathcal{O}=\{o_i\}_{i=1}^{N}$, where $o_i \in \mathcal{O}$ is the $i^{th}$ object instance. This consists of a set $o_i=\{x_i^j\}_{j=1}^{V_i}$ of variable $V_i$ image views, captured from {\it unspecified\/} viewpoints. $x_i^j \in \mathcal{X}$ denotes the $j^{th}$ view of object $o_i$. The goal of LWUMOR is to learn an embedding that supports recognition of new views, objects, and object classes from $\mathcal{O}$. In this work, this is addressed with SSL, defining the surrogate task as object instance classification. Each object instance is treated as a different class, establishing a labelled image dataset $\mathcal{D} = \{(x_i^j, y_i^j) \given y_i^j = i , \forall j \in V_i\}$. The surrogate task is solved by a classifier based on an embedding $f_\theta: \mathcal{X} \rightarrow \mathbb{R}^k$ of parameters $\theta$, which maps image $x$ into $k$-dimensional feature vector $f_\theta(x)$. This is implemented by a convolutional neural network (CNN). It should be emphasized that this surrogate task requires {\it no view alignment or labels.\/} This is unlike previous multiview SSL approaches, which require either view~\cite{shapecode} or camera transformation labels~\cite{egomotion}. 

\subsection{Modeling}\label{sec:modeling}
As is common for CNNs, a classifier can be implemented with a softmax layer
\begin{equation}
P_{Y|X}(i|x) = \frac{\exp({w_i}^T f_\theta(x))}{\sum_{k=1}^{N}\exp({w_k}^T f_\theta(x))},
\label{eq:softmax}
\end{equation}
where $w_i$ is the parameter vector of instance $i$. This is referred as the ``instance classifier" in what follows and is trained by cross-entropy minimization, using the free instance labels for supervision. The optimal embedding and classifier parameters are learned by minimizing the risk
\begin{equation}
    {\cal R} = \sum_{i,j} -\log P_{Y|X}(i|x_i^j)
    \label{eq:risk}
\end{equation}
over the image dataset $\mathcal{D}$.

Even though it is a strong baseline, the softmax classifier is most successful for closed-set classification, where train and test object classes are the same. In general, the learned embedding $f_\theta$ does not have a good metric structure beyond these classes. For this reason, alternative approaches have been more successful for open set problems, such as image retrieval~\cite{image_retrieval}, face identification~\cite{FaceNet,NormFace}, or person re-identification~\cite{ZhengYH16}. These are usually based on metric learning embeddings, such as pairwise~\cite{contrast_loss} or triplet embeddings~\cite{FaceNet}. However, these techniques have problems of their own. Because there are many more example pairs or triplets than single examples in $\mathcal{D}$, they require sampling techniques that are not always easy to implement and lead to slow convergence.

\subsection{Randomization}
In this work, we explore an alternative approach to learn embeddings that generalize beyond the set of training classes, based on the softmax classifier of~(\ref{eq:softmax}). This consists of randomizing the surrogate task and is inspired by previous work in low-shot learning~\cite{SnellSZ17}, where meta-learning techniques re-sample the classes for which the embedding is trained. 
The intuition is that, when the task is changed, the metric structure of the embedding changes as well. This forces the embedding to have a good metric structure over larger regions of the feature space, therefore generalizing better to unseen classes. 
In this work, we consider randomization strategies that leverage the view richness of multiview datasets to achieve better generalization to unseen classes during training. This is critical in the LWUMOR setting, where the goal is to enable the learning of multiview embeddings without dense view datasets or even view labels. We propose to randomize the embedding by using random feature vectors as classifier parameters $w_i$ in (\ref{eq:softmax}). 

The idea is summarized in Fig.~\ref{fig:rand} for a problem where $N = 3$. Fig.~\ref{fig:rand} (a) shows the vectors $w_i$ (solid arrows) learned in feature space with the combination of~(\ref{eq:softmax}) and ~(\ref{eq:risk}). Since cross-entropy minimization only aims to separate the seen instances, this embedding leads to feature distributions such as shown in Fig. \ref{fig:rand} (a). Embeddings of images of the same object are not tightly clustered and can be close to those from other objects. 
A common procedure to encourage better metric structure (in this case Euclidean) is to normalize the embedding to unit norm~\cite{weightimprinting,NormFace,proxynca,FaceNet}, i.e. add a normalization layer at the output of $f_\theta(x)$ such that $||f_\theta(x)||_2=1$. As shown in Fig. \ref{fig:rand} (b), this maps all feature vectors to the unit norm ball. For simplicity, $f_\theta(x)$ is assumed to be normalized in all that follows. 

In this work, we propose to replace the classifier weight $w_i$ by the embedding of a {\it randomly chosen view of object instance $o_i$\/}. This is implemented by defining a {\it view sampler} $\nu_i \in \{1, \ldots, V_i\}$ per 
object instance $i$, which outputs a number between $1$ and $V_i$. This sampler is then used to draw a feature vector $f_\theta(x_i^{\nu_i})$ that serves as the parameter vector $w_i$ of~(\ref{eq:softmax}). The sampled feature vectors are called ``prototypes" for $o_i$, as shown in Fig.~\ref{fig:rand} (c). A softmax temperature parameter $\tau$ is also introduced to control the sharpness of the posterior distribution. 
Larger temperatures originate sharper distributions, smaller temperatures originate more uniform ones.
All these transform the softmax layer into
\begin{equation}
    P^s_{Y|X}(i|x) = \frac{\exp(f_\theta(x_i^{\nu_i})^T f_\theta(x)/\tau)}
    {\sum_{k=1}^{N}\exp(f_\theta(x_i^{\nu_i})^T f_\theta(x)/\tau)},
    \label{eq:prototype}
\end{equation}
where $s=\{f_i^{\nu_i}\}_{i=1}^N$ denotes the set of prototypes used to compute the probability. 

\subsection{Multiview embeddings}

The prototype classifier has the ability to learn a more stable multiview representation than the instance classifier. This, however, depends on the sampling of the prototypes
$f_i^{\nu_i}$ of~(\ref{eq:prototype}). To study the impact of prototype sampling, we consider different {\it randomization schedules\/}, where the view sampler $\nu$ is called more or less frequently during learning, using Algorithm \ref{alg.1}.
\begin{algorithm}[t]
\caption{Randomization schedule}\label{alg.1}
\begin{algorithmic}[1]
\STATE{\textbf{Input} Threshold $t$}
\STATE{Use the view samplers $\nu_i, \forall i$ to select a set of random prototypes ${\cal W} = \{f_\theta(x_1^{\nu_1}), \ldots, f_\theta(x_N^{\nu_N})\}$ to use in~(\ref{eq:prototype}).}\WHILE{Not convergence}
\STATE{ Minimize the risk of (\ref{eq:risk})}
\FORALL{$i \in N$}
\STATE{$u \sim Unif(0,1)$}
\IF{$u < t$}
\STATE{Use $\nu_i$ to resample a new prototype $f_\theta(x_i^{\nu_i})$}
\STATE{$w_i \gets f_\theta(x_i^{\nu_i})$}
\ENDIF
\ENDFOR
\ENDWHILE
\end{algorithmic}
\end{algorithm}
In this algorithm, the threshold $t\in [0,1]$  controls the frequency with which prototypes are changed. If $t=0$, prototypes are fixed, and the embedding is denoted a {\it prototype embedding\/} (PE). If $t=1$, the prototypes can change at every iteration. Fig.\ref{fig:rand} (c-e) illustrates the idea. Starting from an initial prototype set (Fig.\ref{fig:rand} (c)), prototypes are randomized by choosing embeddings of different views of each instance to play the role of prototypes (Fig.\ref{fig:rand} (d-e)) as training progresses. 

Mathematically, prototypes belong to the set $\mathbb{S}=\prod_{i=1}^N \{f_i^j\}_{j=1}^{V_i}$ of all possible combinations of view embeddings across the $N$ object instances. This set has cardinality $|\mathbb{S}| = \prod_{i=1}^N V_i$. The randomization of Algorithm~\ref{alg.1} can thus be seen as replacing~(\ref{eq:softmax}) by an ensemble of $|S|$ classifiers, during training. This is similar to the dropout\cite{Dropout}, but applied to prototypes only. However, unlike dropout, the randomization is structured in the sense that all the prototypes used to replace $w_i$ are embeddings $f_\theta(x_i^{\nu})$ of views from the same object $o_i$. 
This ensembling over views 
makes $f_\theta(x)$ a more stable multiview representation. For this reason, the learned embedding is referred to as a {\it multiview stochastic prototype embedding\/} (MVSPE). 

\subsection{Multiview consistency regularization}$\label{sec:KL}$
The regularization above can be further strengthened by considering the posterior probability distributions of (\ref{eq:softmax}). During training, the feature embedding is guided to move toward the prototypes used at each iteration, as illustrated by the dashed arrows of Fig. \ref{fig:rand} (c-e). This causes the variations in the distributions also shown in the figure. The magnitude of these variations is a measure of the view sensitivity of the embedding. For effective LWUMOR, the feature distributions should not vary significantly with the prototype. This would imply that the different views of the instance were effectively mapped into a view invariant representation. It follows that it should be possible to strengthen the invariance of the embeddings by minimizing these variations, i.e. encouraging the distributions $P_{Y|X}(i|x)$ to remain stable as the set of prototype is varied. This regularization can be enforced by minimizing the average Kullback-Leibler divergence \cite{KLdiv} 
\begin{equation}
L_{KL} = K \sum_{s_p,s_q\in \mathbb{S}, p\neq q}\sum_{k=1}^N P^{s_p}(k|x)\log\left(\frac{P^{s_p}(k|x)}{P^{s_q}(k|x)}\right),
\label{eq:KL}
\end{equation}
where $K=\frac{2}{|\mathbb{S}|(|\mathbb{S}|-1)}$, between all pairs of distributions $P^{s_p}_{Y|X}(i|x)$ and $P^{s_q}_{Y|X}(i|x)$ of prototype sets $s_p$ and $s_q$, where $p\neq q$. When this regularization is used, the resulting embedding is denoted as {\it view invariant stochastic prototype embedding\/} (VISPE). 

\subsection{Scalable Implementation}
In practice, the number of instances in the unlabeled dataset $\mathcal{D}$ can be as large as 30,000. Given the memory capacity of current GPUs, it is impractical to load all object prototypes in memory at each each training iteration. One of the benefits of randomization as a regularization strategy is that it is fully compatible with the sampling of a small subset of object prototypes. In our implementation we use $m=32$ object instances per minibatch. A set ${\cal I} = \{\xi_1, \ldots, \xi_m\}$ of distinct instance indexes is  randomly sampled, defining a subset of view embedding combinations $\mathbb{S}'=\prod_{i=1}^m \{f_{\xi_i}^j\}_{j=1}^{V_{\xi_i}}$ from which prototypes are drawn. Prototype sets are then defined as $s'=\{f_{\xi_i}^{\nu_{\xi_i}}\}_{i=1}^m$ and the posterior  probabilities with
\begin{equation}
    P^{s'}_{Y|X}(\xi_i|x) = \frac{\exp(f_\theta(x_{\xi_i}^{\nu_{\xi_i}})^T f_\theta(x)/\tau)}
    {\sum_{k=1}^{m}\exp(f_\theta(x_{\xi_i}^{\nu_{\xi_i}})^T f_\theta(x)/\tau)},
    \label{eq:prototypeprime}
\end{equation}
where $i \in \{1, \ldots, m\}$.
At each iteration, a pair of prototypes $s'_1$ and $s'_2$ is sampled from the subset of prototype combinations $\mathbb{S}'$, the risk of classifying a training view $x_{\xi_i}^j$ of object instance label $\xi_i$, using prototype set $s'_p$, is computed with
\begin{align}
L_{s'_p}(i,j) &= -\log\left(P^{s'_p}_{Y|X}(\xi_i|x_{\xi_i}^j)\right)
\label{eq:prototype_instance_ensemble_sampled}
\end{align}
for $p\in\{1,2\}$, and the KL divergence with
\begin{equation}
L_{KL'} = \sum_{k=1}^m P^{s'_1}(k|x_{\xi_i}^j)
\log\left(\frac{P^{s'_1}(k|x_{\xi_i}^j)}{P^{s'_2}(k|x_{\xi_i}^j)}\right).
\label{eq:KL_sampled}
\end{equation}
Finally, the stochastic gradient descent (SGD) loss for training example $(x_{\xi_i}^j, {\xi_i}), i \in \{1, \ldots, m\}, j \in \{1, \ldots, V_{\xi_i}\}$ is
\begin{equation}
L= L_{s'_1} + L_{s'_2} + \alpha L_{KL'}.
\label{eq:loss_all}
\end{equation}

In all experiments, we use a temperature $\tau=0.05$ and $\alpha=5$. The implementation is based on Pytorch \cite{pytorch}, using the VGG16 \cite{vgg} model as feature extractor and the output of the last layer as feature vector. A standard SGD was used with learning rate 0.001 to train the network for 300 epochs using batch size of 32.

\section{Experiments}
In this section, we evaluate the different self-supervised learning algorithms on three multiview datasets. 

\begin{figure*}[t] 
 \begin{minipage}{0.6\linewidth}
 \setlength{\tabcolsep}{2.4pt}
\captionof{table} {KNN classification results for various baselines, solving different surrogate tasks. RSPE outperforms all self-supervised learning methods, VGG16 pretrained model and instance classifiers.} 
\vspace{-2pt}
\begin{tabular}{|c|c|c|c|c|c|c|c|}
        \hline
        Datasets & Surrogate & \multicolumn{2}{c|}{ModelNet} & \multicolumn{2}{c|}{ShapeNet}
         & \multicolumn{2}{c|}{ModelNet-S}\\
        Methods / Classes & Task & seen & unseen & seen & unseen & seen & unseen \\
        \hline
        Chance & N/A & 3.3 & 10.0 & 3.3 & 4.0 & 3.3 & 10.0\\
        \hline
        \hline
        Pretrained \cite{vgg} & N/A & 62.7 & 52.7 & 63.9 & 58.1 & 58.2 & 55.2\\
        \hline
        \hline
        Autoencoder\cite{autoencoder} & Context & 31.8 & 37.2 & 29.8 & 26.3 & 34.7 & 38.8\\
        Egomotion \cite{egomotion} & Motion  & 32.4 & 34.7 & 72.6 & 47.1 & 33.0 & 35.2\\
        Puzzle \cite{puzzle} & Sequence & 34.4 & 41.5 & 67.8 & 48.6 & 34.8 & 42.4\\
        UEL \cite{UEL} & Data Aug. & 47.9 & 46.5 & 68.7 & 53.4 & 46.4 & 48.2 \\
        ShapeCode \cite{shapecode} & View & 39.4 & 46.5 & 67.1 & 42.3 & 38.8 & 47.2\\
        MVCNN\cite{MVCNN} & N/A & 39.6 & 48.1 & 30.3 & 32.4 & 36.7 & 44.8\\
        Triplet\cite{FaceNet} & N/A & 70.1 & 62.4 & 81.2 & 61.2 & 64.7 & 62.1\\ 
        Instance classifier & N/A & 57.7 & 58.9 & 69.3 & 60.4 & 52.3 & 54.6\\
        \hline
        \hline
        PE & Object & 69.7 & 61.7 & 81.6 & 63.8 & 62.1 & 60.4\\
        MVSPE & Object & 70.3 & 63.2 & 82.4 & 64.6 & 64.6 &  62.1\\
        VISPE & Object & \textbf{71.2} & \textbf{64.4} & \textbf{82.9} & \textbf{65.5} & \textbf{66.2} & \textbf{64.3}\\
        \hline
    \end{tabular}
\label{tab:classification}
\vspace{-6pt}
\end{minipage}
\begin{minipage}{0.37\linewidth}
\centering
     \begin{tabular}{ccc}
      \includegraphics[width=0.35\linewidth]{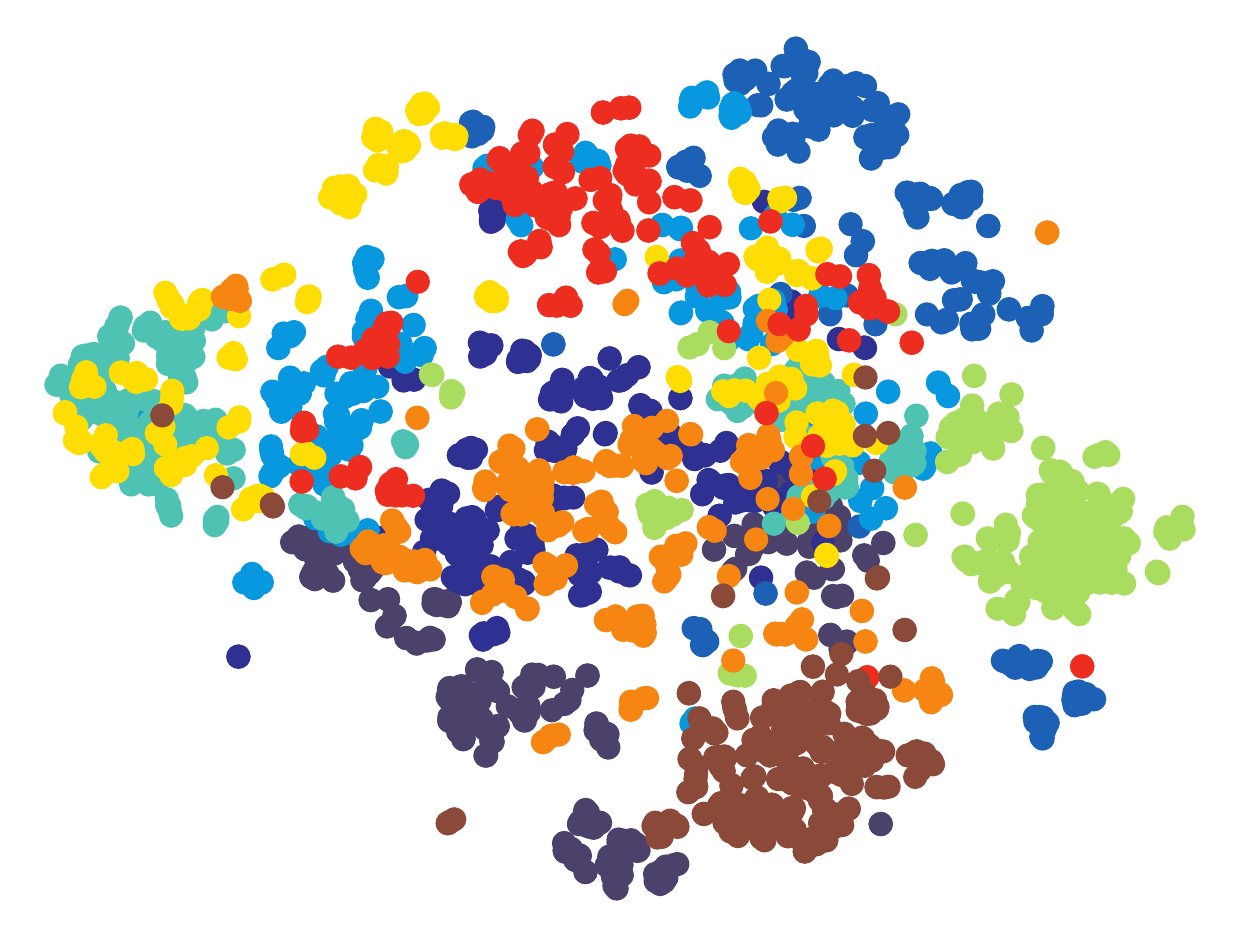}  
      & \includegraphics[width=0.35\linewidth]{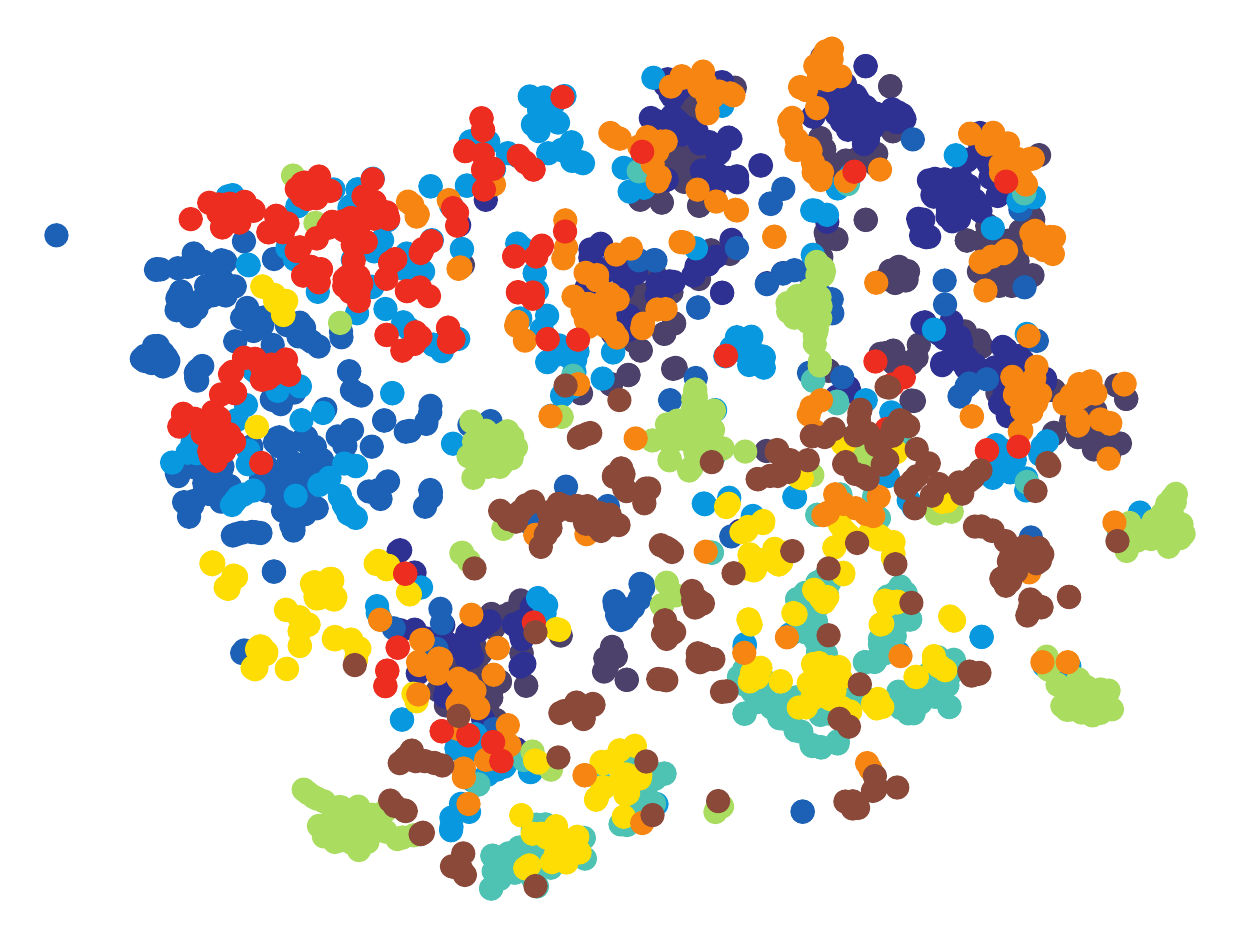}  \\
      \footnotesize{Pretrained} &  \footnotesize{ShapeCode} \\
      \includegraphics[width=0.35\linewidth]{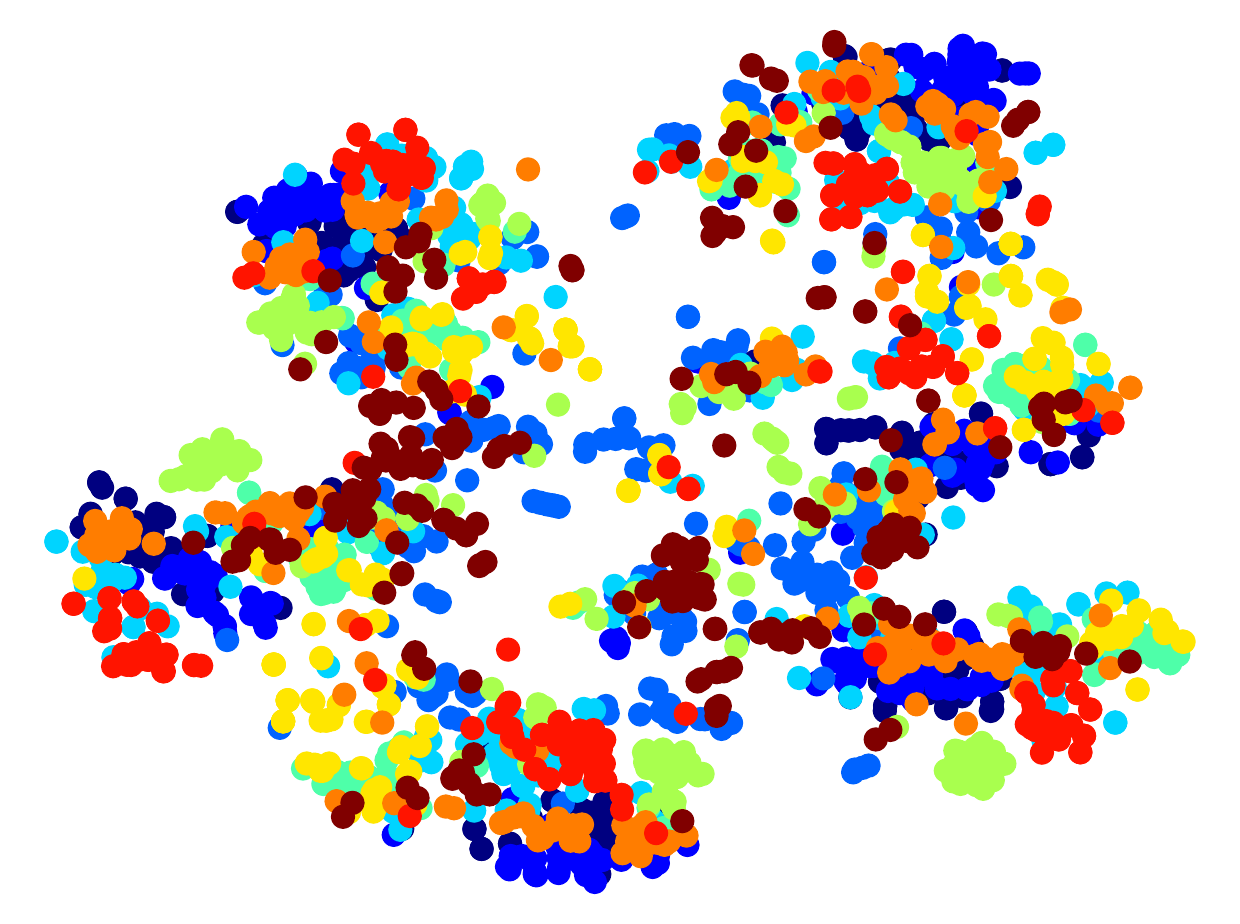}  
      & \includegraphics[width=0.35\linewidth]{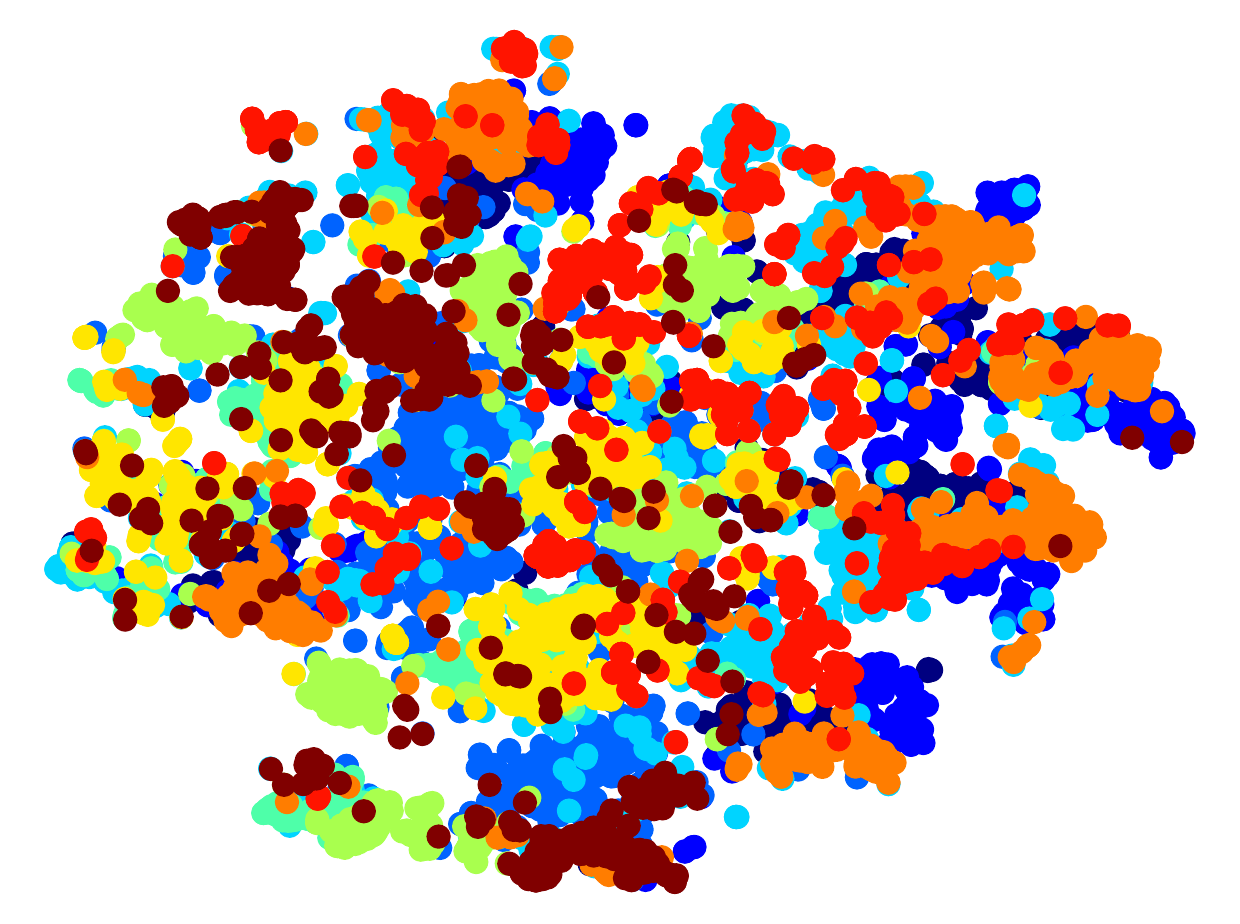} \\
      \footnotesize{Egomotion} &  \footnotesize{Puzzle}\\
      \includegraphics[width=0.35\linewidth]{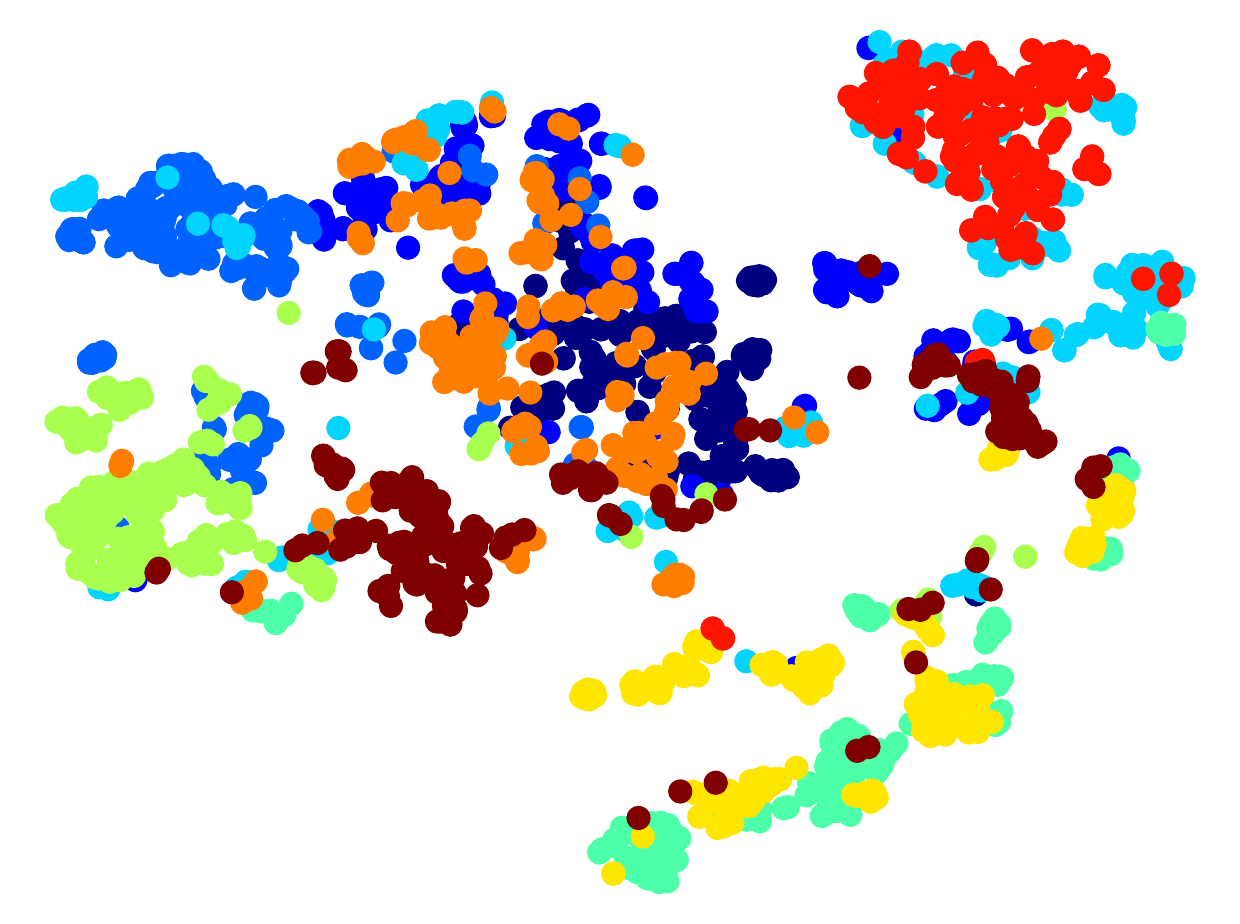}
      & \includegraphics[width=0.35\linewidth]{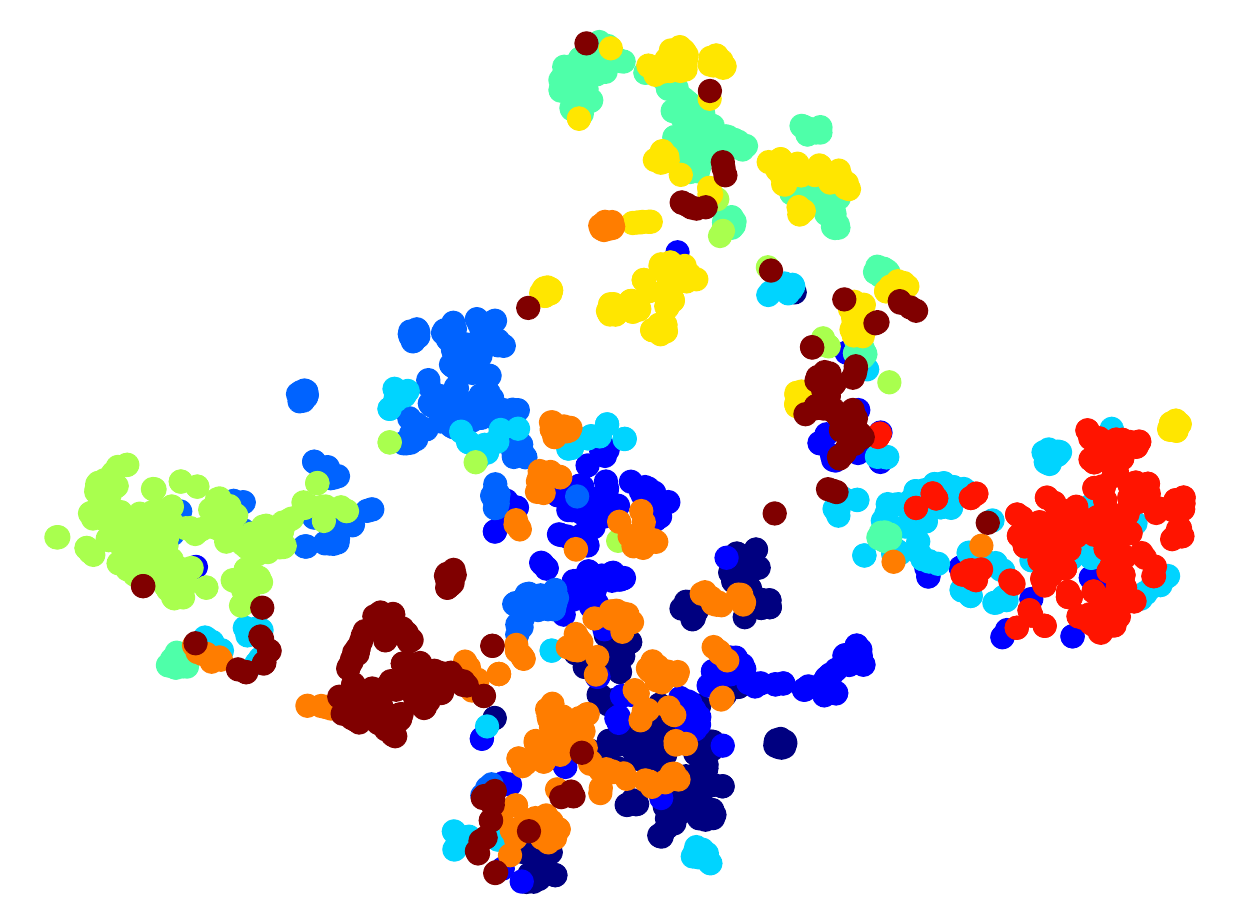} \\
      \footnotesize{Triplet} &  \footnotesize{VISPE} 
     \end{tabular}
     \captionof{figure}{TSNE visualization of \textbf{unseen class} embeddings. Each color represents a class. RSPE produces more structured embedding.}
     \label{fig:tsne_vis}
     \vspace{-6pt}
 \end{minipage}

\end{figure*}

\subsection{Dataset}
Three datasets, Modelnet40\cite{modelnet}, Shapenet\cite{shapenet} and ModelNet-S, were used in all experiments. Instead of the rendering process of \cite{shapecode}\footnote{We do not have access to the rendered images.}, we adopted the rendering approach and dataset\footnote{https://github.com/suhangpro/mvcnn} widely used in the multiview literature\cite{MVCNN,GVCNN,Kanezaki16,AngularTC,TC,Ho_2019_CVPR}. Given a synthetic CAD model, 12 views are rendered around it at every 30 degrees. The virtual camera elevates 30 degrees and points to the center of the model. Please see \cite{MVCNN} for more details.

\noindent\textbf{Modelnet}\cite{modelnet} is a synthetic dataset of 3,183 CAD models from 40 object classes. We follow the seen/unseen class split of \cite{shapecode}, where unseen classes are those of Modelnet10, a subset of Modelnet40. The standard training and testing partitions~\cite{MVCNN,GVCNN,Kanezaki16,Ho_2019_CVPR} are adopted.

\noindent\textbf{ShapeNet}\cite{shapenet} is a synthetic dataset 
of 55 categories following the Wordnet\cite{wordnet} hierarchy. We use the rendered images from \cite{MVCNN}, which contains 35,764 training objects and 5,159 test objects. The seen/unseen class split procedure is identical to \cite{shapecode}, using the 30 largest categories as seen and the remaining 25 as unseen classes.  

\noindent\textbf{Modelnet-S} is sampled by ourselves to resemble a dataset with missing views. This is a subset of Modelnet and shares its train/test setup as well as seen/unseen classes.   

\subsection{Baselines}
We consider SSL baselines that solve different surrogate tasks, ranging from context, to motion, view, data augmentation and sequence based, as discussed in Section \ref{sec:related_work}. All baselines except Jigsaw puzzle\cite{puzzle} use the same backbone (VGG16\cite{vgg}) and features are extracted from the last network layer. All methods are trained from scratch. For \cite{puzzle}, we refer to the original paper for architecture details and the use of pool5 feature. The implementation of these baselines is discussed below and more details can be found in the supplementary materials.

\noindent\textbf{Pretrained} \cite{vgg}  is a VGG16 model pre-trained on ImageNet\cite{imagenet_cvpr09} using class supervision. 

\noindent\textbf{Autoencoder} \cite{autoencoder} is trained to reconstruct the input image, 
using an L2 loss to measure the difference between input and reconstruction.

\noindent\textbf{Egomotion} \cite{egomotion} predicts the camera motion between 2 images. Given a pair of images, features are extracted, concatenated and fed into stacked fully connected layers to predict the relative view point difference. Assuming only $V$ viewpoints exist in the dataset, the model will output $V-1$ probabilities, corresponding to the $V-1$ viewpoints differences. For architecture details see supplementary material.

\noindent\textbf{Jigsaw puzzle} \cite{puzzle} crops 9 patches from the $ 255\times 255$ input images and shuffles them. The surrogate task is to solve the puzzle. Based on the public source code\footnote{\scriptsize{\url{https://github.com/bbrattoli/JigsawPuzzlePytorch}}}. 

\noindent\textbf{UEL} \cite{UEL} treats each image as a class and learns a data augmentation invariant feature. Based on the author's code\footnote{\scriptsize{\url{https://github.com/mangye16/Unsupervised_Embedding_Learning}}}.

\noindent\textbf{ShapeCode}\cite{shapecode} 
reconstructs the subsequent views given an object view. We use the loss function proposed in the original paper to train the network. To accommodate the different rendering conditions, the network inputs and generates $224 \times 224$ images instead of $32 \times 32$. 

\noindent\textbf{MVCNN}\cite{MVCNN} inputs all views of an object and averages their feature vectors, feeding the result to a fully connected classifier that predicts the object identity. 

\noindent\textbf{Triplet}\cite{FaceNet} is a metric learning approach that learns from triplets of examples: an anchor (input) image, a positive image (from the same object as the anchor) and a negative image (from a different object). Margin $1$ performed best in this setting.

\noindent\textbf{Instance classifier} treats each object as a class and trains a VGG16 classifier to minimize (\ref{eq:risk}).

\subsection{Classification}
All baselines are tested on the 3 datasets, using no labels for training. Inference is based on $k$ nearest neighbor classification, where $k$ is the number of images of the class with fewest objects in the dataset. This is $960$, $468$ and $500$ for ModelNet, Shapenet and ModelNet-S, respectively. Each experiment is repeated for seen and unseen classes per dataset.  

\begin{figure}
    \begin{tabular}{cc}
    \includegraphics[width=0.46\linewidth]{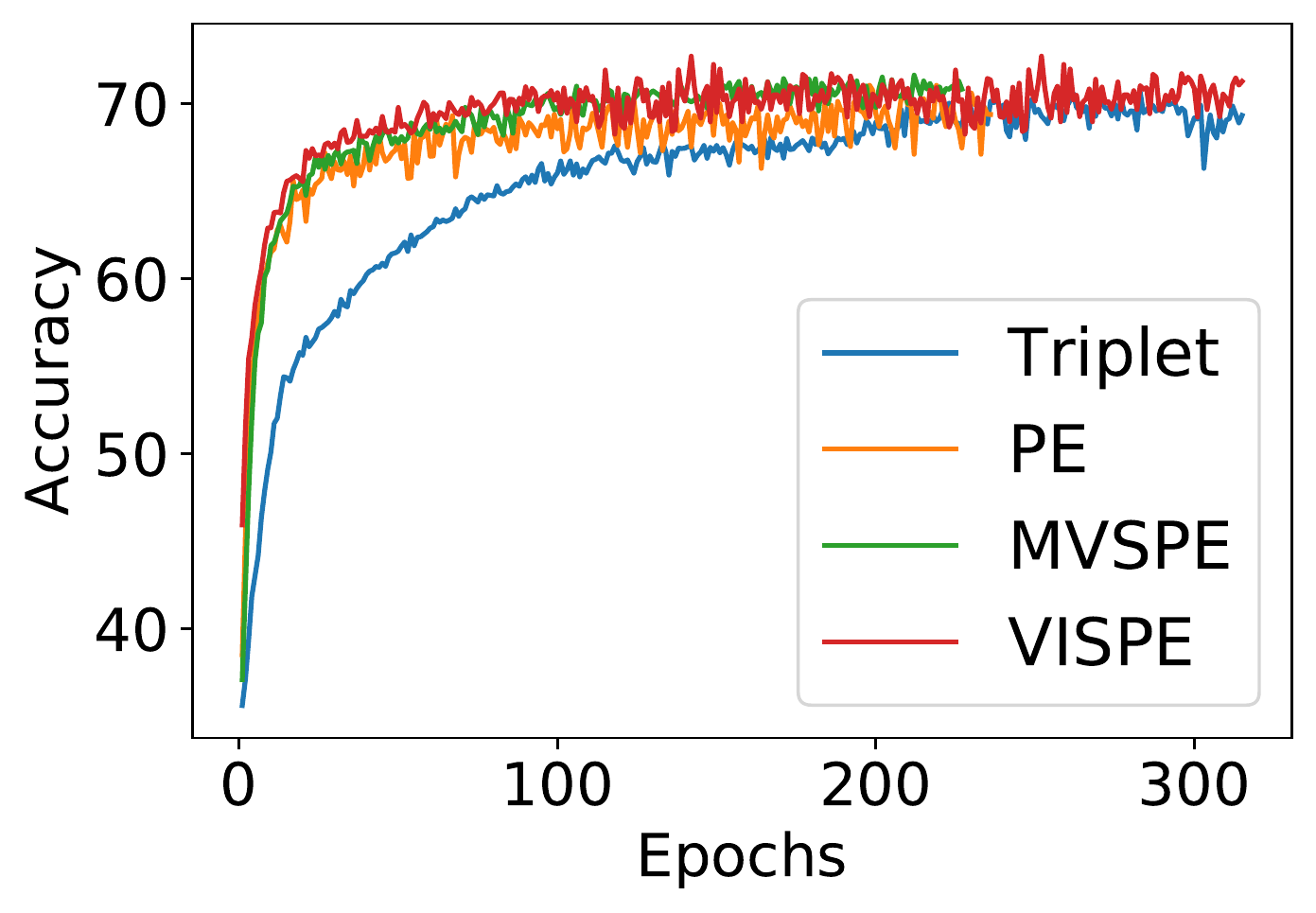}
    &\includegraphics[width=0.46\linewidth]{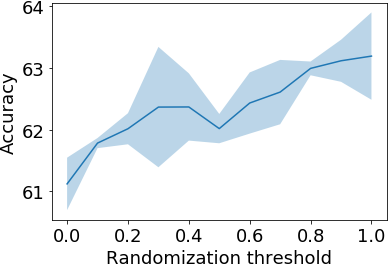}\\
    \footnotesize{(a)} & \footnotesize{(b)}
    \end{tabular} 
    \vspace{-3pt}
    \caption{(a) Convergence rate of proposed methods and triplet loss. (b) Effect of different randomization threshold on unseen class accuracy.}
    \label{fig:converge_and_rand_thres}
    \vspace{-6pt}
\end{figure}

Table~\ref{tab:classification} shows that all previous surrogate tasks perform poorly for MV-SSL. All proposed methods outperform all baselines, regardless of surrogate task. The only competitive baselines are methods that distinguish objects: instance classifier and triplet embedding. The triplet loss has results comparable to MVSPE but, as discussed in Sec.~\ref{sec:modeling} and shown in Fig. \ref{fig:converge_and_rand_thres} (a), much slower training convergence. While all proposed methods converge in around 80 training epochs for ModelNet, it requires more than 200. 
Overall, VISPE has the best performance in all datasets, for both seen and unseen classes. This shows that the surrogate task of learning {\it object invariants\/} leads to more robust SSL for multiview data.
Fig. \ref{fig:converge_and_rand_thres} (b) shows the effect of the randomization threshold of Algorithm~\ref{alg.1}, presenting the average accuracy on unseen classes over ten experiments per threshold. Despite the variance of these results, it is clear that randomization during training strengthens model generalization to unseen classes.

\subsection{Retrieval and Clustering}
Ideally, the learned embedding should map images from the same class close together and images from different classes apart, \textit{even for \textbf{unseen} classes}. To test this, Kmeans\cite{kmean} is used to cluster the image embeddings of unseen classes. Two metrics are used to evaluate clustering quality: recall @ K and normalized mutual information (NMI)\cite{NMI}. NMI is defined as $\frac{2I(A,C)}{H(A),H(C)}$,
where $I$ denotes mutual information, $H$ entropy, $A=\{a_1, \dots, a_n\}$ where $a_i$ is the set of images assigned to class $i$, and  $C=\{c_1, \dots, c_n\}$,  where $c_j$ is the set of images of ground truth class $j$.  Both metrics are popular in the metric learning literature\cite{proxynca,SongXJS15,NIPS2016_6200}.

Table~\ref{tab:recall_nmi_low_shot} shows results for Modelnet. Again, triplet  is the only baseline competitive  with the proposed embeddings, although weaker, and VISPE clearly achieves the best performance.
Its effectiveness is highlighted by the large NMI gains. The tightness of its clusters can also be verified in Fig. \ref{fig:tsne_vis}, which presents a TSNE visualization of the feature embeddings. Note how VISPE clustering better separates the different colors, which identify the different object classes.

\subsection{Few-shot object recognition}
The generalization strength of the different embeddings is further tested by experiments with few shot classification. Table~\ref{tab:recall_nmi_low_shot} shows classification accuracy of unseen classes when $k$ images are labeled per object class. A linear SVM is trained on the labelled feature vectors of Modelnet~\cite{modelnet} unseen classes, and used to classify its test set. Similarly to the previous experiments, only the triplet embedding is competitive with MVSPE and VISPE, and VISPE achieves the best performance.

\subsection{Dependence on number of objects and views}
We next consider how many views and objects are needed to learn an embedding that generalizes to unseen classes. To study this, we sample a subset of ModelNet objects and a subset of views per object. The VISPE embedding is then trained on the sampled data and tested on the unseen classes. 
Fig.~\ref{fig:object_view_num_and_labelornot} (a) shows that classification accuracy saturates around 40 objects per class and 8 views per object. Interestingly, when the number of objects is small, capturing more views per object compensates for the lack of object diversity. This is of importance for applications, since it suggests that the embedding could be quickly retrained on a relatively small set of objects, e.g. when a robot has to be deployed on a totally new environment. 

\begin{table}
\small
 \setlength{\tabcolsep}{2.5pt}
\captionof{table} {Left: Recall @ k and NMI on Modelnet unseen classes. Right: low shot accuracy for k labeled images.} 
    \begin{tabular}{|c|c|c|c|c|c||c|c|c|}
        \hline
        & \multicolumn{5}{|c||}{Retrieval and Clustering} 
        & \multicolumn{3}{|c|}{Low shot} \\
         & \multicolumn{4}{|c|}{Recall} & &
         \multicolumn{3}{|c|}{k Images}\\
        Methods & @1  & @2  & @4  & @8 & NMI 
        & 1 & 3 & 5\\
        \hline
        \hline
        Pretrained \cite{vgg}  & 94.5 & 96.6 & 98.2 & \textbf{99.3}  & 46.7 
        & 34.3 & 46.8 & 51.2  \\
        \hline
        Autoencoder\cite{autoencoder}   & 81.7 & 86.8 & 92.3 & 95.2 & 25.4 
        & 25.0 & 30.0 & 28.0 \\
        Egomotion \cite{egomotion}   & 73.4 & 80.7 & 88.0 & 92.9 & 7.5
        & 15.1 & 18.1 & 19.9\\
        Puzzle \cite{puzzle}   & 77.8 & 84.1 & 89.8 & 94.0 & 21.9
        & 21.1 & 26.8 & 29.3\\
        UEL \cite{UEL}   & 77.8 & 85.4 & 91.6 & 95.7 & 24.6 & 23.8 & 30.9 & 34.2\\
        ShapeCode \cite{shapecode}    & 83.4 & 88.5 & 93.4 & 96.2 & 27.4 & 28.8 & 36.1 & 39.5  \\
        MVCNN \cite{MVCNN}   &  80.3 & 86.7 & 91.7 & 95.0 & 19.3 & 21.6 & 27.0  & 29.5\\
        Triplet \cite{FaceNet} & 90.8 & 94.7 & 97.4 & 98.8 & 48.2 & 41.4 & 50.3 & 54.5\\
        Instance classifier & 89.1 & 92.5 & 95.6 & 97.4 &  37.1 & 28.3 & 42.2 & 48.4 \\
        \hline
        \hline
        PE  & 91.2 & 95.0 & 97.2 & 98.5 &  48.2 & 40.2 & 49.7 & 52.9  \\
        MVSPE & 92.4 & 95.4 & 97.7 & 98.9 &  48.4 & 41.5 & 50.8 & 54.2 \\
        VISPE & \textbf{95.5}  & \textbf{97.7} & \textbf{98.6 }& 99.2 &  \textbf{51.1} & \textbf{43.1} & \textbf{52.5} & \textbf{55.9}\\
        \hline
    \end{tabular}
    \vspace{-6pt}
\label{tab:recall_nmi_low_shot}
\end{table}

\subsection{Trade-off of training with labels}
\begin{figure}
    \begin{tabular}{cc}
    \includegraphics[width=0.46\linewidth]{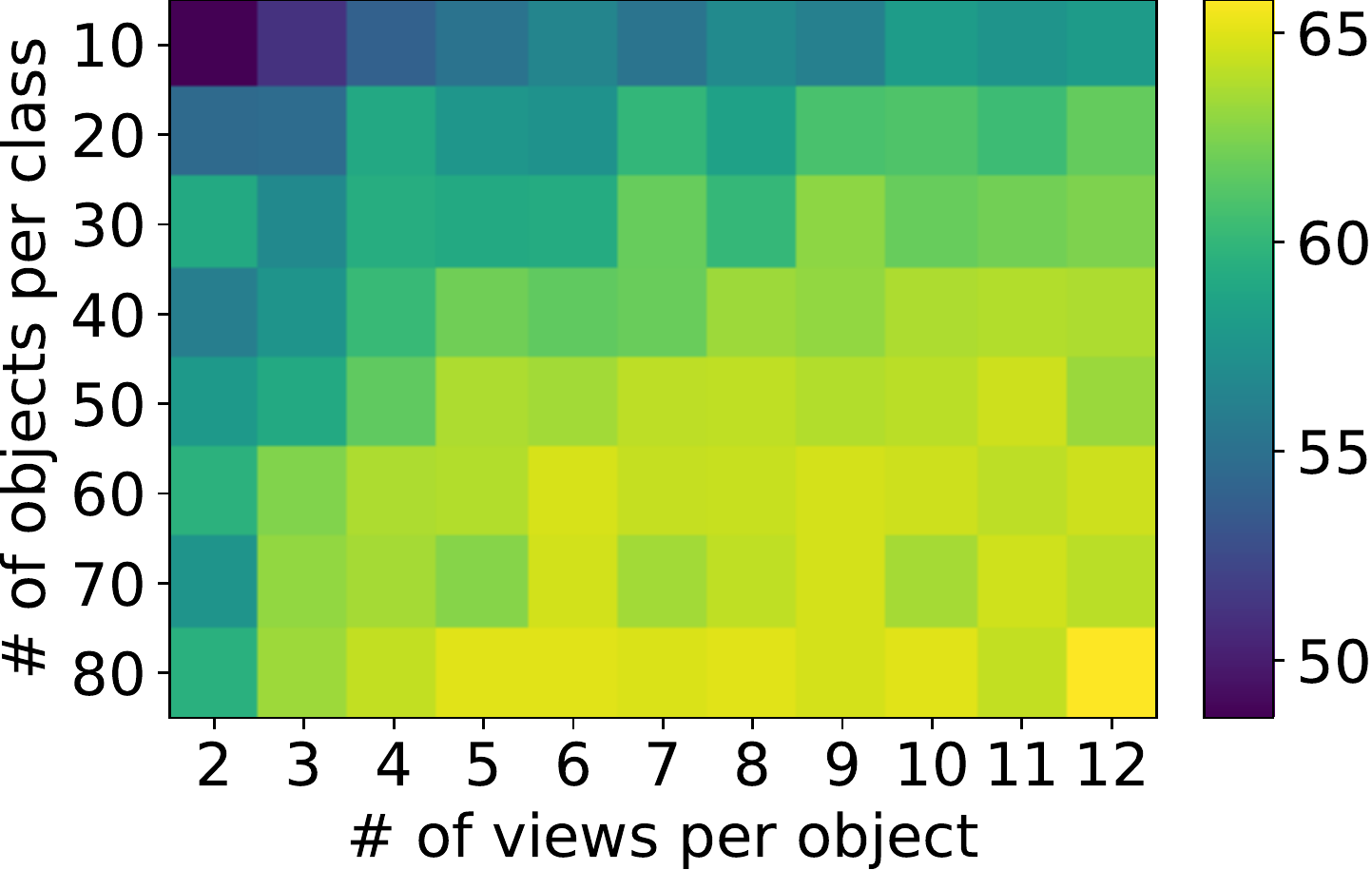}
    &\includegraphics[width=0.46\linewidth]{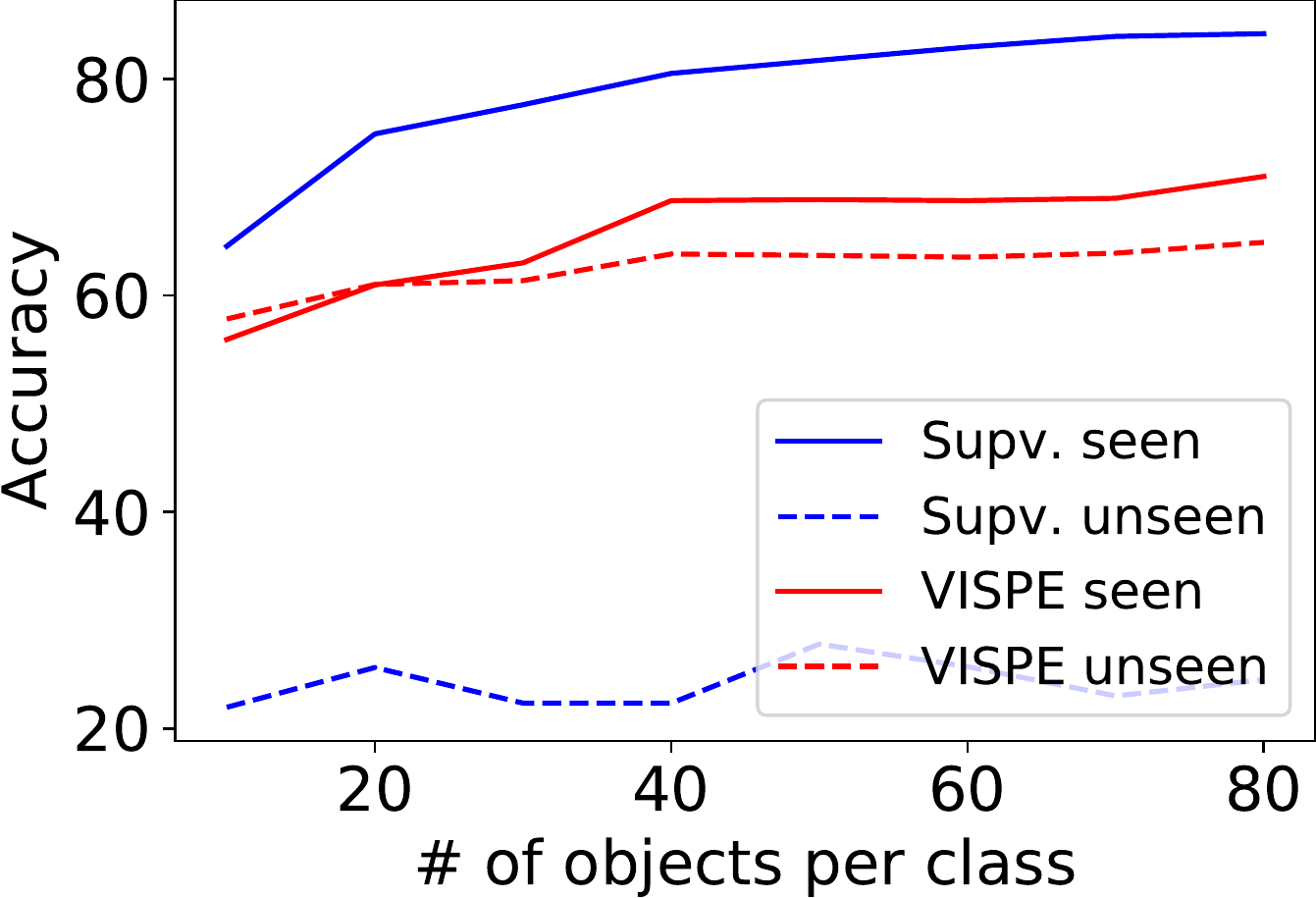}\\
    \footnotesize{(a)} & \footnotesize{(b)}
    \end{tabular}
    \vspace{-5pt}
    \caption{ (a) Accuracy (represented by color) of VISPE on unseen classes, as a function of views per object and objects per class in training set. (b) Trade-off of training with labels as a function of object per class.}
     \label{fig:object_view_num_and_labelornot}
     \vspace{-12pt}
\end{figure}

Even though supervised learning requires more labeling effort, it performs better on predefined classes. For example, training VGG16\cite{vgg} with labels on seen classes of ModelNet and ShapeNet yields $84.1\%$ and $87.5\%$ on its test set. However, the performance of KNN on unseen classes drops significantly to $20.9\%$ and $35.1\%$, which is much worse than most SSL results in Table~\ref{tab:classification}. This begs the question of when SSL should be used. As shown in Fig. \ref{fig:object_view_num_and_labelornot} (b), when few objects per class are available for ModelNet, VISPE is a better choice than supervised learning, because it generalizes much better ($>30\%$) on unseen classes and maintains comparable performance ($<10\%$) on seen classes.


\section{Conclusion}
In this work, we made several contributions to MV-SSL. We started by discussing the current impractical assumption of fully supervised multiview recognition, which requires intensive labeling. We then relaxed this assumption by investigating MV-SSL methods, where only ``free labels" (image to object association) are required. Embeddings that generalize to both seen and unseen data were then learned with variants of this MV-SSL surrogate task. These variants differ in the regularization  used to encourage object invariant representations. We started by leveraging view information by choosing the embedding of a random object view as the object prototype. A randomization schedule was then proposed to sample prototypes stochastically. This can be seen as an ensembling over views, to encourage stable multiview embeddings. To strengthen the learning of object invariants, we finally proposed a multiview consistency constraint. The combination of all these contributions produced a new class of view invariant stochastic prototype embeddings (VISPE). These embeddings were  shown to outperform other SSL methods on seen and unseen data for both multiview classification and retrieval. While we have not studied the semi-supervised setting, where few labels are provided, in great detail, this setting is also supported by VISPE. 
We believe that these are important contributions for the much needed extension of multiview recognition to the LWUMOR setting of Figure~\ref{fig:intro_fig}, which is of interest for many real world applications.

\vspace{2pt}
\noindent \textbf{Acknowledgments} This work was partially funded by NSF awards IIS-1637941, IIS-1924937, and NVIDIA GPU donations.

\clearpage
{\small
\bibliographystyle{ieee_fullname}
\bibliography{ref}
}
\end{document}